\documentclass[10pt,twocolumn,letterpaper]{article}
\pdfoutput=1

\usepackage[breaklinks=true,bookmarks=false]{hyperref}

\usepackage{eso-pic}
\usepackage{xspace}

\newif\ifcvprfinal
\cvprfinalfalse
\def\cvprfinalcopy{\global\cvprfinaltrue}

\def\@maketitle
{
	\newpage
	\null
	\vskip .375in
	\begin{center}
		{\Large \bf \@title \par}
		\vspace*{24pt}
		{
			\large
			\lineskip .5em
			\begin{tabular}[t]{c}
				\ifcvprfinal\@author\else Anonymous CVPR submission\\
				\vspace*{1pt}\\%This space will need to be here in the final copy, so don't squeeze it out for the review copy.
				Paper ID \cvprPaperID \fi
			\end{tabular}
			\par
		}
		\vskip .5em
		\vspace*{12pt}
	\end{center}
}

\def\abstract
{%
	\centerline{\large\bf Abstract}%
	\vspace*{12pt}%
	\it%
}

\def\affiliation#1{\gdef\@affiliation{#1}} \gdef\@affiliation{}

\long\def\@makecaption#1#2{
	\setbox\@tempboxa\hbox{\small \noindent #1.~#2}
	\setlength{\@ctmp}{\hsize}
	\addtolength{\@ctmp}{-\@figindent}\addtolength{\@ctmp}{-\@figindent}
	\ifdim \wd\@tempboxa >\@ctmp
	{\small #1.~#2\par}
	\else
	\hbox to\hsize{\hfil\box\@tempboxa\hfil}
	\fi}

\makeatletter
\DeclareRobustCommand\onedot{\futurelet\@let@token\@onedot}
\def\@onedot{\ifx\@let@token.\else.\null\fi\xspace}

\makeatother

\usepackage{amsmath}
\usepackage{times}
\usepackage{epsfig}
\usepackage{graphicx} %,subfigure}
\usepackage[export]{adjustbox}
\usepackage{amssymb}
\usepackage{booktabs}
\usepackage{caption}
\usepackage{subcaption}
\usepackage[algoruled,boxed,lined]{algorithm2e}

\cvprfinalcopy

\newcommand{\scoresimbol}{s}
\newcommand{\labelsimbol}{l}

\newcommand{\TAM}[1] {}
\newcommand{\NOA}[1] {}
\begin{document}

%%%%%%%%% TITLE
\title{Inner-Scene Similarities as a Contextual Cue for Object Detection}
\date{}
\author{
Noa Arbel\\ {\tt\small noaarl@cs.technion.ac.il} \\
\and
Tamar Avraham\\ {\tt\small tammya@cs.technion.ac.il}\\
\and
Michael Lindenbaum\\ {\tt\small mic@cs.technion.ac.il}\\
%Technion, Israel Institute of Technology\\
}

\maketitle

%\thispagestyle{empty}
%%%%%%%%% ABSTRACT
% Arxiv gets up to  1920 characters abstract
%Abstract from thesis
\begin{abstract}
%%%%%%%%% ABSTRACT
Using image context is an effective approach for improving object detection. Previously proposed methods used contextual cues that rely on semantic or spatial information. In this work, we explore a different kind of contextual information: inner-scene similarity. We present the CISS (Context by Inner Scene Similarity) algorithm, which is based on the observation that two visually similar sub-image patches are likely to share semantic identities, especially when both appear in the same image. CISS uses {\em base-scores} provided by a base detector and performs as a post-detection stage. For each candidate sub-image (denoted {\em anchor}), the CISS algorithm finds a few similar sub-images (denoted {\em supporters}), and, using them, calculates a new enhanced score for the anchor. This is done by utilizing the base-scores of the supporters and a pre-trained dependency model. The new scores are modeled as a linear function of the base scores of the anchor and the supporters and is estimated using a minimum mean square error optimization. This approach results in: (a) improved detection of partly occluded objects (when there are similar non-occluded objects in the scene), and (b) fewer false alarms (when the base detector mistakenly classifies a background patch as an object). 
This work relates to Duncan and Humphreys' "similarity theory," a psychophysical study. which suggested that the human visual system perceptually groups similar image regions and that the classification of one region is affected by the estimated identity of the other. Experimental results demonstrate the enhancement of a base detector's scores on the PASCAL VOC dataset.
\end{abstract}
%%%%%%%%% BODY TEXT
\section{Introduction}

\begin{figure}[t]
	\centering
	\begin{subfigure}[b]{0.2\textwidth}
			\includegraphics[width=1.1\textwidth]{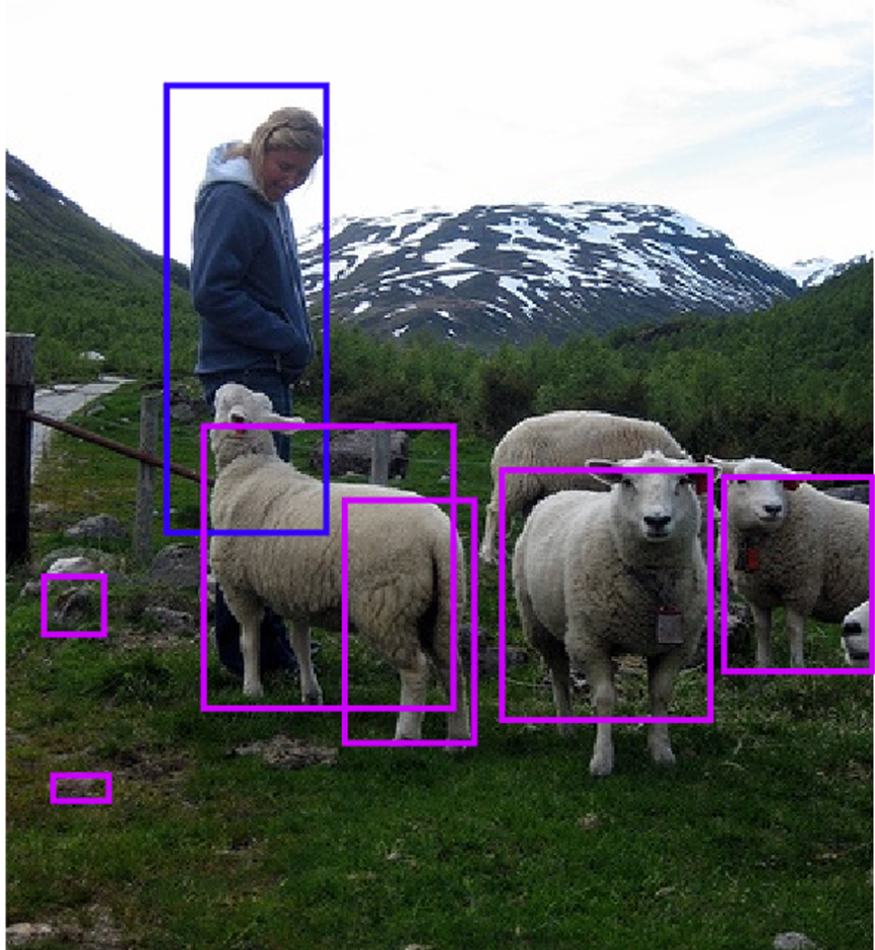}
		\caption{Baseline}
		\label{fig:pic_overview_base}
	\end{subfigure}%
	\qquad
	\begin{subfigure}[b]{0.2\textwidth}
			\includegraphics[width=1.1\textwidth]{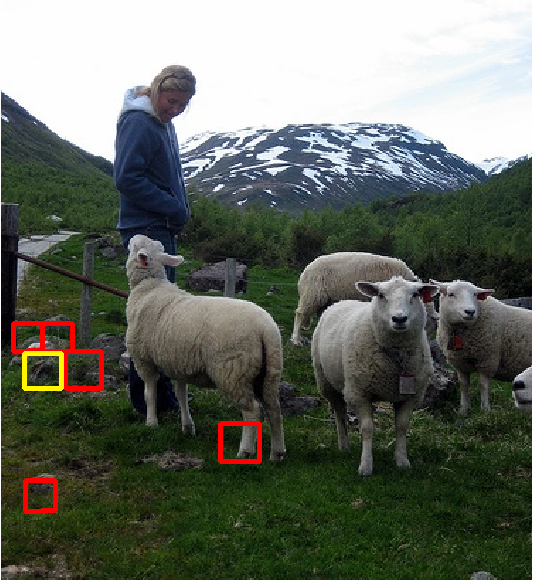}			
	\caption{False positive detection}
	\label{fig:pic_overview_example1}
	\end{subfigure}%
	\vfill
		\begin{subfigure}[b]{0.2\textwidth}
				\includegraphics[width=1.1\textwidth]{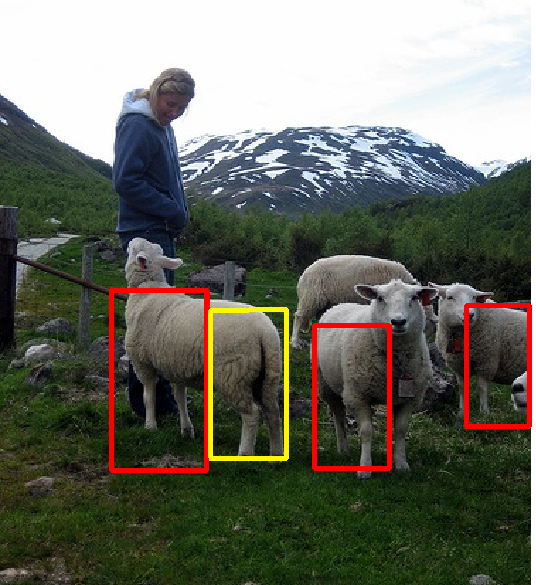}
			\caption{False positive detection}
			\label{fig:pic_overview_example2}
		\end{subfigure}% 			
	\qquad
	\begin{subfigure}[b]{0.2\textwidth}
			\includegraphics[width=1.1\textwidth]{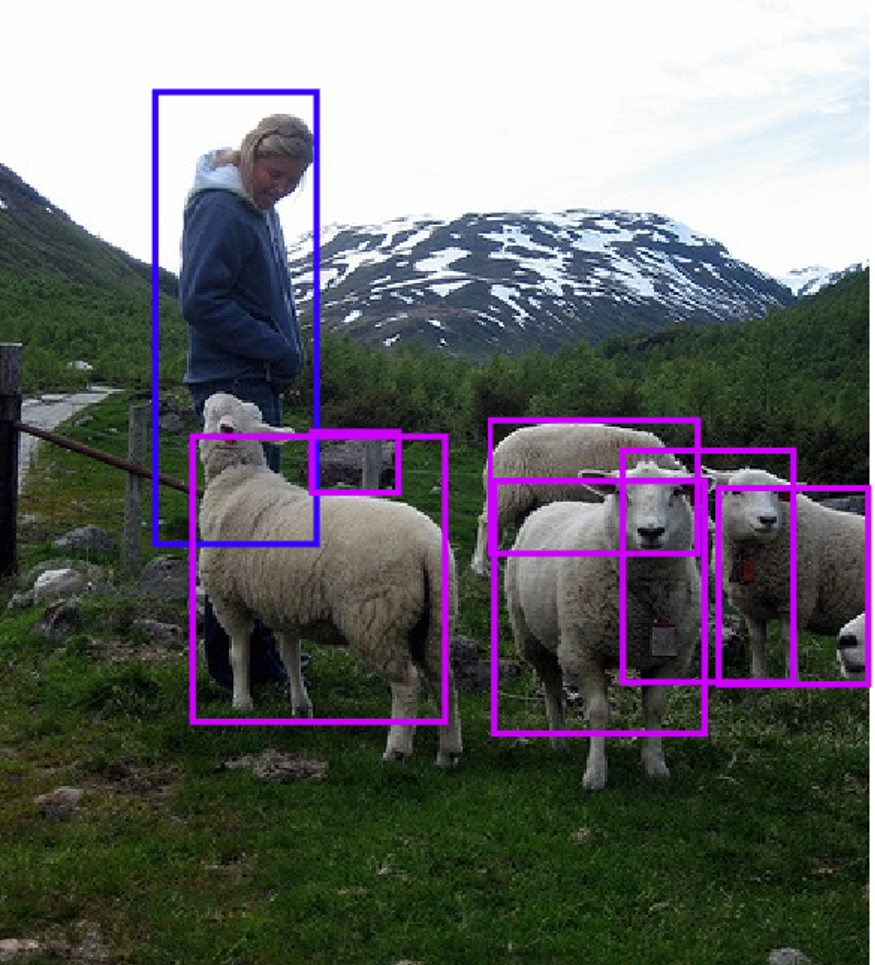}
		\caption{Final CISS results}
		\label{fig:pic_overview_CISS_res}
	\end{subfigure}%
	\caption{Overview of the proposed CISS algorithm.\\		 
		(a) Shows the 7 top scoring image regions by a base detector (Fast-RCNN). In blue is the best detection for the `person' category and in pink are the best 6 detections for the `sheep' category; (b) and (c) show two false positive detections in yellow, and their similar patches in red; and (d) shows the 7 top scoring image regions by CISS.}
	\label{fig:pic_overview}
\end{figure}

% detection
Object detection is composed of object localization and object classification, and is one of the major computer vision challenges. 
Traditional object detection and categorization used tailored appearance features to recognize objects in images. In recent years, appearance features are automatically generated when training deep convolutional neural networks (CNNs) for classification or segmentation tasks.
Until recently, it was common to train a classifier using a set of annotated images containing single objects, and to afterwards apply the classifier on sub-images of full scenes. The sub-images were usually extracted by a naive sliding window method, or by segmentation-based methods that extracted `proposals' (candidate object locations). Some recent CNNs were designed to infer both the regions of interest and their detection scores using one forward pass~\cite{redmon2015you,fasterRCNNren2015}.

% context
Several models that use contextual information for improving detection performance have been proposed. The most common context-based methods involve semantic cues and spatial cues. Some of those methods use semantic context by applying statistics on various object class co-occurrences in training data~\cite{choi2012tree}, while other methods model the probability of an object class to appear in the image given the image's gist~\cite{gist2006}. Some methods use spatial cues by utilizing the statistics of the locations, or relative locations, for object classes in scenes~\cite{mottaghirole}.

% motivation
This work explores a different kind of contextual information: inner-scene similarity. While intra-image patch similarities are used for other image processing and computer vision applications such as super-resolution~\cite{avraham2011ultrawide,glasner2009super}, texture classification~\cite{malik2001contour,varma2005statistical}, segmentation and object search~\cite{avraham2006attention}, information of this type has not been explored as a cue for improving object detection accuracy. We show that by utilizing inner-scene similarity, we can sometimes overcome two main difficulties that current object detectors do not handle very well: (1) detection in the presence of background clutter and (2) detection of partly-occluded objects. Moreover, we show that the cue of inner-scene similarity can reduce false alarms.

% CISS main idea
We present the CISS (Context by Inner Scene Similarity) algorithm.
CISS is based on the observation that two visually similar sub-image patches are likely to share semantic identities, especially when both appear in the same image.
A similar observation was used in~\cite{avraham2006attention}, which followed psychophysical studies~\cite{duncan1989visual} and proposed a dynamic visual search attentional framework driven mainly by inner-scene similarity. 

% how the algorithm works
CISS uses `base-scores' provided by a base detector and performs as a post-detection stage, proposing enhanced scores. It is generic in the sense that it does not depend on the detector used and can enhance any detector's scores. In this work, we demonstrate CISS using Fast R-CNN~\cite{girshickICCV15fastrcnn} as the base detector.

We follow~\cite{avraham2006attention} and use the minimum mean square error linear estimator (MMSE) to calculate the new scores. However, unlike~\cite{avraham2006attention}, which considered the dependencies between similarity and identity, we quantify (a) the dependencies between sub-image similarity and the correlation of the two corresponding base-scores, and (b) the dependencies between the base-score of one candidate object and the identity of another candidate, given the similarity between them. 
The statistics of these dependencies are learned from training data and utilized by the estimation process.
We evaluate our algorithm on the well-known PASCAL VOC benchmark~\cite{everingham2010pascal}.

%and the newer MS-COCO \cite{lin2014microsoft}. While the PASCAL VOC is more common for testing object detection methods, the latter dataset is more suitable for evaluating our achievements as it contains richer contextual information.\\
%\subsection{Contribution}

% Figure 1 description
Figure~\ref{fig:pic_overview} demonstrates CISS's contribution. In Figure~\ref{fig:pic_overview_base} and~\ref{fig:pic_overview_CISS_res} we show 7 image regions for which Fast R-CNN~\cite{girshickICCV15fastrcnn} provided the highest base-score, and the 7 top scoring regions by CISS, respectively. In Figure~\ref{fig:pic_overview_example2} and~\ref{fig:pic_overview_example1} we show two example anchors (in yellow) with the similar `supporters' found by CISS (in red). Both examples depict false alarms of the base detector: in Figure~\ref{fig:pic_overview_example1} the false alarm is a background error, and in Figure~\ref{fig:pic_overview_example2} the false alarm is a "localization error"~\cite{hoiem2012diagnosing}.

% invariant vs non-invariant features. What CISS can do
Object recognition and object detection methods usually aim to utilize features that are invariant to viewing conditions and are scene independent. In this work we wish to do the contrary: we exploit the fact that two objects (or background patches) that share an identity and appear in the same image are usually similar also in features that are non-invariant to viewing conditions such as weather, luminance, texture, and even pose. Moreover, given that both objects belong to the same class, they will usually belong to a common subclass, and will have more in common than any two objects sharing identity but captured at different times and locations.
Therefore, while detectors recognize objects mainly on the basis of those parts that are meaningful for characterizing the objects' classes, CISS uses a complementary similarity measure based on color and texture, and only weakly on the spatial pixel arrangement.
%By using a similarity measure that introduces features different from those used by the local detector, we avoid repeating information that was already exploited by the detector.
As a result, CISS can increase the score of an object, given that other similar objects in the scene were assigned a higher base-score, and can decrease the score of a background patch given that other background patches with similar texture and color were assigned very low base-scores. It can therefore lead to better detection of objects of low visibility, or partly occluded objects, and can reduce false alarms.

The rest of this paper is organized as follows. In Section 2 we review related work. We describe the CISS algorithm in Section 3 and report experimental results in Section 4. Section 5 concludes the paper.

\section{Related Work}

\subsection{Traditional Object Detection}
Advances in object detection (and other vision tasks) during the previous decade have relied mainly on the use of SIFT~\cite{sift2004lowe} and HOG features~\cite{hog2005dalal}. Two dominant approaches that use those features are the Bag of Visual Words~\cite{csurka2004bow} and the Deformable Parts Model (DPM)~\cite{FelzenszwalbDPM}.
The DPM represents objects using mixtures of a root and deformable part HOG filter, forming a 'star' model. The detector obtains scores for (almost) all locations in the image using a sliding window. On top of this model, the authors created a mixture of star models for each category. This detector was a leading detection method for some time~\cite{everingham2010pascal}, until neural network based image classifiers broke through computation and scalability barriers~\cite{imagenetClassification2012}.

\subsection{CNN Based Detectors}
One of the first high performing deep CNN based object detectors is R-CNN~\cite{girshick14CVPR}. It takes  numerous  candidate  object locations, called "proposals"~\cite{selective2013uijlings}, and classifies each using a forward pass of an object recognition network such as the one in~\cite{imagenetClassification2012}.
To achieve better running time and more accurate localization results, the author of~\cite{girshickICCV15fastrcnn} created Fast R-CNN, a single-stage training algorithm that jointly learns to classify object proposals and refine their spatial locations. One forward pass is used to estimate all classes and locations of the candidates in the image.

Faster and more current works make region proposals a part of the networks. In~\cite{fasterRCNNren2015}, one network implements end-to-end detection, including region extraction. The YOLO model~\cite{redmon2015you} also performs detection by a single forward pass. It predicts both bounding boxes and class probabilities that are associated with entries in a fixed image grid.

\subsection{Object Detection and Context}
Several models that use contextual information to improve detection performance have been proposed~\cite{2010contextsurvey}. The most common contextual cues involve semantic context and spatial context. The former take into account other object classes that may appear in the image and use statistics on co-occurrences in training data. The latter learn the expected locations for object classes in scenes and use them as priors for detection.
 
Contextual cues can also be categorized in other ways: those that use global context~\cite{gist2006} vs. those that use local context, or those that use object-scene interaction vs. those that use object-object interaction~\cite{desai2011discriminative}. The contextual cues may be integrated as a post-processing step given the detector output~\cite{choi2012tree, rabinovich2007objects}, or integrated within the detector itself~\cite{mottaghirole}. A good example of an object detection method that uses context can be found in the work of Choi et al.~\cite{choi2012tree}. The authors propose a graphical tree model that captures the contextual cues of co-occurrences and spatial relationships. They show that their model can improve the results of object detection tasks (both for object localization and for presence prediction), as well as other scene understanding tasks that local detectors alone cannot solve, such as detecting  objects that are out-of-context.

CNNs that are trained for the task of image classification do not learn context implicitly. Lately there have been some attempts to use local context cues in CNNs by modifying the architecture to include some candidate surroundings~\cite{gidaris2015object}, or by enlarging the input box of the candidates~\cite{zhu2015segdeepm}

\subsection{Similarity Between Images or Image Patches}
Traditionally, color and texture attributes are used to determine the similarity between images (for tasks such as texture classification~\cite{malik2001contour, varma2005statistical}) or between image parts (for image segmentation~\cite{liu2013robust, selective2013uijlings}, denoising~\cite{dabov2007image}, image completion and image extrapolation~\cite{criminisi2004region, kwok2010fast}, super-resolution~\cite{avraham2011ultrawide, glasner2009super}, texture synthesis~\cite{de1997multiresolution} and saliency detection~\cite{avraham2010esaliency, goferman2012context}).
It is common to describe each image patch by some feature vector and to then measure similarities by simple distance functions (i.e., Euclidean), or by histogram intersection (when the feature vectors are histograms). 
The features may be based on texture (e.g, be  SIFT-like~\cite{liu2010exploring} or on responses of basis filters~\cite{varma2005statistical}) or on color information~\cite{arvis2011generalization}.
A well-tested approach to describe patch texture is the bag of textons proposed by Malik et al.~\cite{malik2001contour} and used for texture segmentation.

Several CNN based works have managed to perform well using 'Siamese' networks. The 'Siamese' network acts as a feature extractor after which a comparison method can be applied. In~\cite{simo2015discriminative}, a simple $L_2$ distance metric is used to compare patches. In~\cite{zagoruyko2015learning} and~\cite{matchnet2015han} an additional network is learned for the comparison stage. In MatchNet~\cite{matchnet2015han}, the full network is disassembled during prediction, to allow faster and more efficient computations.

Patch texture similarity can also be measured by applying a material classification network on the patches and then comparing their output (we adopted this method in our work; see details in Section 3). Bell et al.~\cite{bell2015material} proposed a material classifier that uses large texture patches for training. 
%Other work by Cimpoi et al.~\cite{textureswild2014} aims to enhance texture classification by using appearance attributes (such as stripes or polka-dots).
\section{The Algorithm}

\subsection{Overview and Notations}
A general description of CISS is given in Algorithm~\ref{algo:CISS}.
CISS starts by applying a {\em base detector} on the input image. The base detector outputs a set of triplets $\{a_i = (b_i, c_i, s_i)\}$, where $b_i$ is a rectangle and $s_i$ is a score that quantifies the likelihood of the sub-image $I[b_i]$ to depict an object of the category $c_i$. We refer to $s_i$ as the {\em base-score}, and to each $a_i$ with a base-score higher than a threshold as an {\em anchor}.

For each anchor, CISS finds similar rectangular image regions $\{I[b_{i,j}]\}$. We refer to these regions as {\em supporters}. The visual dissimilarity between an anchor and its supporters is measured by a function $d$ and denoted $d_{i,j}=d(I[b_i], I[b_{i,j}])$. How CISS finds the supporters is explained in detail in Section~\ref{find-similar-patches}. 

For each supporter, CISS obtains the base-score ${s_{i,j}}$ associated with the likelihood of $I[b_{ij}]$ to depict an object of category $c_i$. The score of the anchor is then revised, taking into account $s_{i}$, $\{s_{i,j}\}$, $\{d_{i,j}\}$, and the dependency functions $\gamma_{ss}$ and $\gamma_{ls}$. 
%The revised score $p_i$, is obtained by a minimum mean square error estimator (MMSE) that works on the assumption that $p_i$ is a linear function of $s_{i}$ and $\{s_{i,j}\}$. 
We assume that the revised score $p_i$ is a linear function of $s_{i}$ and $\{s_{i,j}\}$. Using this assumption, $p_i$ is obtained by a minimum mean square estimator (MMSE).
For details on the estimation process see Section~\ref{re-evaluate-scores}.
% algorithm
\begin{algorithm}[t]
\begin{minipage}{0.4\textwidth}
	\SetKwInput{KwData}{Input}
	\SetKwInput{KwResult}{Output}
	\KwData{\begin{itemize} \item image $I$ \item a set of anchors $\{a_i = (b_i, c_i, s_i)\}$, described each by a box, a hypothesized category, and a base-score.
			\end{itemize}}
	\ForEach{anchor $a_i$}{
		\begin{itemize}
			 \item Find similar supporters $\{b_{i,j}\}$, with appearance distances $d_{i,j}=d(I[b_i], I[b_{i,j}])$ (as described in Section~\ref{find-similar-patches}) 	
			 \item Obtain base-scores for the supporters ${s_{i,j}}$	
			 \item Re-estimate the anchor score using $\{s_{i}\}$, $\{s_{i,j}\}$, $\{d_{i,j}\}$. %, and the dependency functions $\gamma_{ss}$ and $\gamma_{ls}$ (as described in Section~\ref{re-evaluate-scores})
		\end{itemize}
	}
	\KwResult{New estimated scores $\{p_i\}$\\ }
	\caption{The CISS algorithm}\label{algo:CISS}
	\end{minipage}
\end{algorithm}

\subsection{Finding Similar Supporters}\label{find-similar-patches}
%Descrition of the similarity measure
To quantify the appearance similarity of two patches (anchor and a potential supporter), we extract color and texture descriptors from each.
For color, we use an HSV color histogram with 10 bins per channel.
To describe the texture of a patch we use a pre-trained neural network based material classifier~\cite{bell2015material}. 
This classifier provides 23 confidence values corresponding to 23 materials, for every image pixel. We describe the textural characteristics of a sub-image patch with a $23$-dimensional vector, by averaging the activation maps over the patch's pixels.

We then compute the chi-square distances between the associated color and textural descriptors, and compute a weighted average between them:
\begin{equation}\label{eq:dist}
d_{i,j} = \alpha\cdot(\chi^2_{\text color})_{i,j} + \beta\cdot(\chi^2_{\text texture})_{i,j}~~~.
\end{equation}
In addition, we add a pyramid based comparison in order to roughly compare the spatial arrangement inside the patches.

Naturally, other descriptors and similarity measures can be used. We experimented also with histograms of textons \cite{malik2001contour} and the results were only slightly inferior.

\begin{figure*}[pt]
	\centering
	\begin{subfigure}[b]{0.29\textwidth}
			\includegraphics[width=\linewidth]{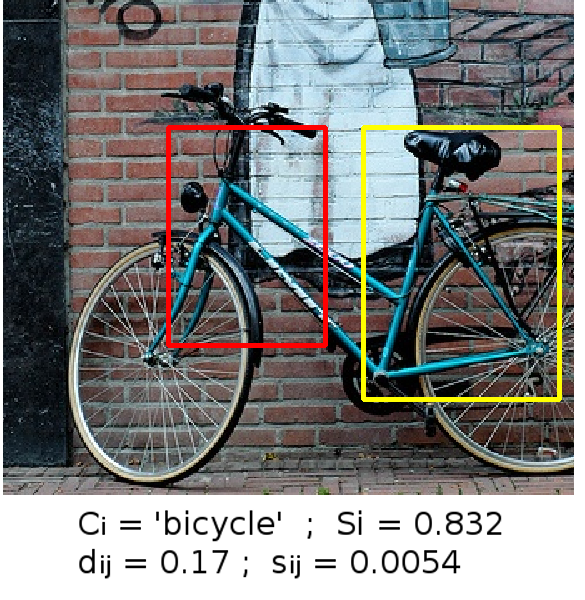}			
			\caption{}
			\end{subfigure}
			\hfill %\qquad
			\begin{subfigure}[b]{0.42\textwidth}
					\includegraphics[width=\linewidth]{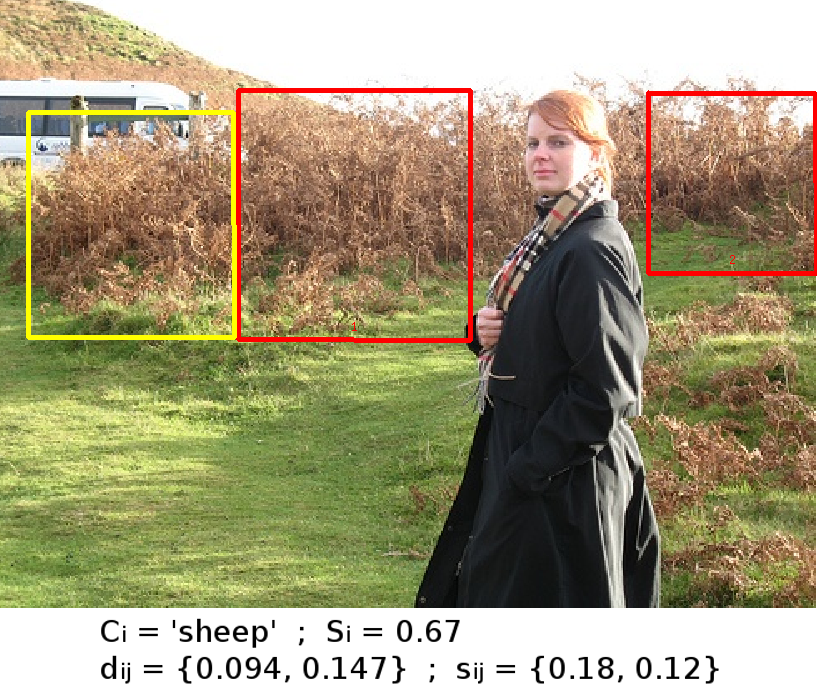}			
				\caption{}
			\end{subfigure}
				\hfill %\qquad
			\begin{subfigure}[b]{0.25\textwidth}
				\includegraphics[width=\linewidth]{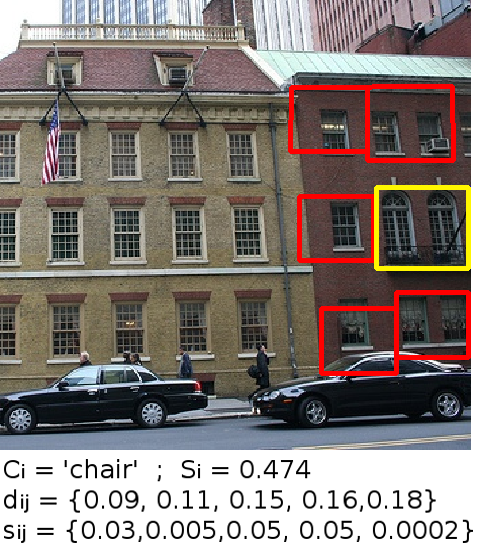}
				\caption{}
			\end{subfigure}
	\caption{Patch similarity results. In yellow, the anchor detection; in red, the supporting windows. Base-scores and appearance distances are reported below each image.}
	\label{fig:similarity_examples}
\end{figure*}

% short description of the way we use the similarity measue and a refrance to the figure
For every candidate anchor we iteratively choose the most similar supporting region which is of the same size as the anchor (up to 20\% difference) and does not overlap with previously chosen supporters. This is done efficiently as described in Section \ref{implementaiton-details}. 
Some examples of the results of this procedure can be viewed in Figure~\ref{fig:similarity_examples}.

\subsection{Re-evaluating Anchor Scores}\label{re-evaluate-scores}
\subsubsection{Estimating scores with the linear MMSE}\label{MMSE}
In the CISS approach, we consider each object identity $l_i$ as a binary random variable ($l_i = 1$ when $I[b_i]$ depicts an object from category $c_i$ and $l_i = 0$ otherwise) and consider $s_i$, and $\{s_{i,j}\}$, ~$j=1,...,n_{\text sup}^i$, the base-scores, as continuous random variables. Once we have selected the $n_{\text sup}^{i}$ supporters for the anchor $a_i$, we can re-estimate its score, or its probability to belong to the class $c_i$.
We model $p_i$, the estimate of $l_i$, as a linear function of $s_{i}$ and $\{s_{i,j}\}$,
\begin{equation}\label{eq:estimation}
p_i =  m_0 + m_i \cdot s_i + \sum\limits_{j=1}^{n_{\text sup}^{i}}m_{i,j} \cdot s_{i,j}~~~.
\end{equation}
Using the common technique of MMSE estimation~\cite{papoulis2002}, we then calculate
\begin{equation} \label{eq:MSE_solution}
p_i = E[l_i]+\vec{m^*} (\vec{\scoresimbol}-E(\vec{\scoresimbol}))~~~,
\end{equation}
where $\vec{s}=(s_i, s_{i,1},...,s_{i,n_{\text sup}^{i}})$, $E(s_i)$ and $E(s_{ij})$ are set to the expected value of the category's base-score for a random patch. $E[l_i]$ is the probability for a random patch to be associated with the category, and $\vec{m^*}$ are the coefficients for which the square error has a minimum.
The value of the coefficients is given by:
\begin{equation}\label{eq:coefficient solution}
\vec{m^*} = C^{-1} \cdot R~~~,
\end{equation}
where $C$ includes the covariances between $s_i$ and $\{s_{i,j}\}$ in the first row and column, and the covariances between the $s_{i,j}$'s in rows and columns 2 to $(n_{\text sup}^{i} + 1)$. R is a vector consisting of the covariances between $s_i$ and $l_i$ and between each $s_{i,j}$ and $l_i$.

\subsubsection{Characterizing the dependency behavior} \label{characterize-dependency-behavior}
$C$ and $R$ are estimated using training data. We model the covariance between two base-scores and between a base-score and the $l_i$ identity as a descending function of the feature-space distance between the associated sub-images, and denote it with $\gamma$. 

\begin{equation*}
\gamma_{s,s}(d_{i,j}) \triangleq \tilde{cov}(s_i,s_{i,j})
\end{equation*}
\begin{equation*}
\gamma_{s,s}(d(I(b_{i,j_1}),I(b_{i,j_2}))) \triangleq \tilde{cov}(s_{i,j_1},s_{i,j_2})
\end{equation*}
\begin{equation*}
\gamma_{l,s}(d_{i,j}) \triangleq \tilde{cov}(l_i,s_j)~~~.
\end{equation*}

Where $\tilde{cov}(s_i,s_{i,j})$, $\tilde{cov}(s_{i,j_1},s_{i,j_2})$ and $\tilde{cov}(l_i,s_j)$ are functions that best fit the covariance values of the training data.
% $\gamma_{s,s}$ and $\gamma_{l,s}$ from the statistic of the training images. 
As demonstrated in Section~\ref{ex_Characterize_Dependency_Behavior}, we extract sub-images with known, manually annotated identities and obtain the base-scores for them. Then, using a large set of such sub-image pairs (with base-scores $s_{k_1}$ and $s_{k_2}$, and identities $l_{k_1}$ and $l_{k_2}$), we calculate the covariances values for each distance interval:

\begin{equation} \label{eq:gamma_scores}
cov_{s,s}(d) = E[s_{k_1}s_{k_2}](d)-E[s_{k_1}](d)E[s_{k_2}](d)
\end{equation}
and
\begin{equation} \label{eq:gamma-score-and-label}
%\gamma_{l,s}(d) = 
cov_{l,s}(d) = E[l_{k_1}s_{k_2}](d)-E[l_{k_1}](d)E[s_{k_2}](d).
\end{equation}

Then, we compute exponential descending functions $\gamma_{s,s}$ and $\gamma_{l,s}$  that are the best fit for $cov_{s,s}$ and $cov_{l,s}$.

\section{Experiments}
In Section~\ref{ex_Characterize_Dependency_Behavior} we describe in detail the estimation of the dependency characteristics (the $\gamma$ functions). In Section~\ref{implementaiton-details} we discuss some CISS implementation issues and in Section~\ref{VOC results} report results for the VOC PASCAL detection challenge. In Section~\ref{discussion} we further discuss the advantages and weaknesses of CISS.
%Section \ref{Dependency Behaviour} details the estimation of the $\gamma$ functions. Section \ref{VOC results} shows results and implementation details of CISS on PASCAL dataset and section \ref{discussion} discuss the results.

\subsection{Characterization of the Dependency Behavior}\label{ex_Characterize_Dependency_Behavior}

%\paragraph{Patches Collection}\label{Patches}
% intro. for the dataset
The dependency functions between base-scores of two sub-images and between a base-score and sub-image identity, are inferred from training data. For this task we used the SUN09 database~\cite{xiao2010sun}. We chose this dataset as it contains many fully annotated images (including the annotations of background regions). We extracted patches from the first 3000 images of the training set. The collected patches are either the bounding boxes of objects, or rectangular patches containing background. Background patches are either boxes contained inside a background segment (and then described by one background category), or randomly selected patches that may be associated with multiple background categories. The number of collected background patches roughly equaled the number of objects in the image.
For each selected patch we obtained the base-score provided by Fast R-CNN~\cite{girshickICCV15fastrcnn}.
For each pair of patches coming from the same image, we measured the similarity using the distance described in equation~\eqref{eq:dist} when $\alpha=\beta=0.5$.
Each pair was classified as a `same' pair if both patches are similarly annotated, or as a `not-same' pair otherwise. 

%\paragraph{Texton Creation}\label{Texton}
%For each image in the set, compute the filter response for all the sets of filters checked in \cite{varma2005statistical} densely (we may consider to use only the MR8 set). Then, we'll create a dictionary of K visual words with k-means clustering (we can consider using other methods like Expectation maximization clustering \cite{yu2012bag}).
%\\
%Choosing K: in the papers, we found that for each texture class 10 \cite{varma2005statistical} to 30 \cite{alvarez2012texton} words are learned (30 was for 3 color channels). So, we can chose K as 10 times the average number of segments in the images.
%\paragraph{Texture Map Creation}\label{minc model}
%The work of \cite{bell2015material}, present two stage image material segmentation. We use the first CNN that gives a probability value for a 256*256 patch to be one of 23 material categories. Then, combining the materials output from different location and image scale using the same CNN converted to a sliding window. The results is 23 heat maps of the size of the image, as seen in \cite{bell2015material} pipeline.
%
%\paragraph{Color Information}\label{Color}
%To add color feature, the patches will be converted to HSV space and quantized to 10 values per channel.
%\paragraph{Combined distance}\label{final distance} For final measure, we use the chi square distance measure as detailed in previous section. We checked result for variation of weights between color and texture. For the next experiments, we used weights is 0.5 for color and 0.5 for texture.\\
%

\begin{figure}[h]
	\centering
	\includegraphics[width=0.43\textwidth,height = 5.2cm]{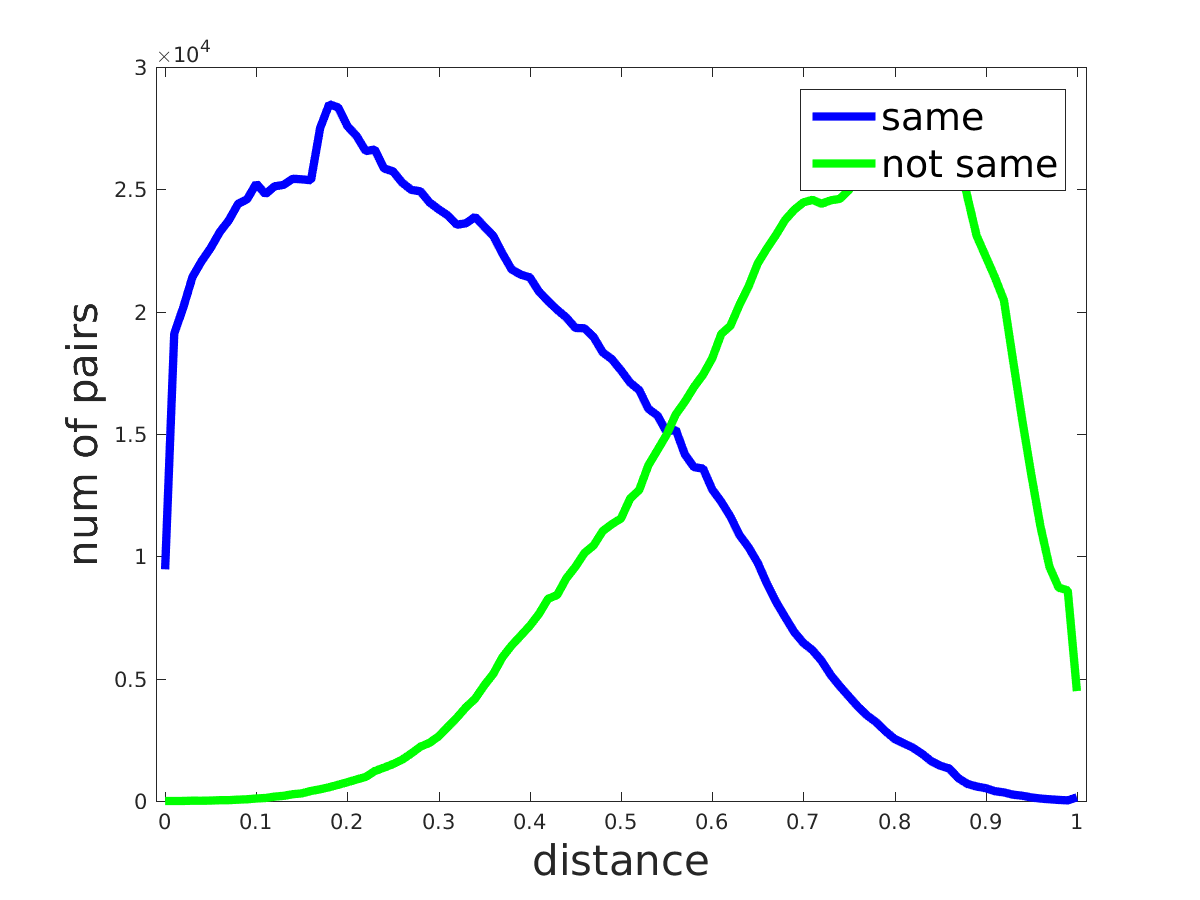}
	%\caption{Histogram of distances for equal texture and color weights}		
	\caption{Distribution of patch pair distances}
	\label{fig: Same vs Not Same Statistics fig}
\end{figure}

Figure~\ref{fig: Same vs Not Same Statistics fig} shows the distribution of the distances associated with `same' pairs and the distribution of distances associated with `not-same' pairs. As expected, we see a strong correlation between inner-scene similarity and the probability that the two patches share or don't share an identity. Specifically, we see that if the measured visual distance between two patches is low (below 0.25), there is a very high probability that they describe objects (or background) from the same class.

%      It present that the similarity measure distinguishes well between pairs of patches of the same class and pairs of patches that do not belong to the same class. The result also shows that if a pair of patches has a low distance value, it is almost certain that the patches are from the same class. For the CISS experiment, we consider supporters to be similar to their anchor if the distance between them satisfies:  $\distancesimbol(\anchorsimbol,j) \leq 0.25$,

\begin{figure}[tb]
	%\centering
	\begin{subfigure}[t]{0.05\textwidth}
	(a)
	\end{subfigure}
	\begin{subfigure}[t]{0.36\textwidth}{}
		\includegraphics[width=\textwidth,height=4cm,valign=t]	{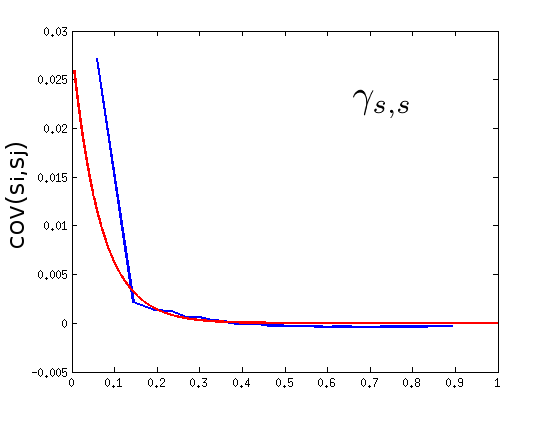}
		%\caption{}
	\end{subfigure}%
	\\
	\begin{subfigure}[t]{0.05\textwidth}
	(b)
	\end{subfigure}
	\begin{subfigure}[b]{0.36\textwidth}
		\includegraphics[width=\textwidth,height=4cm,valign=t]	{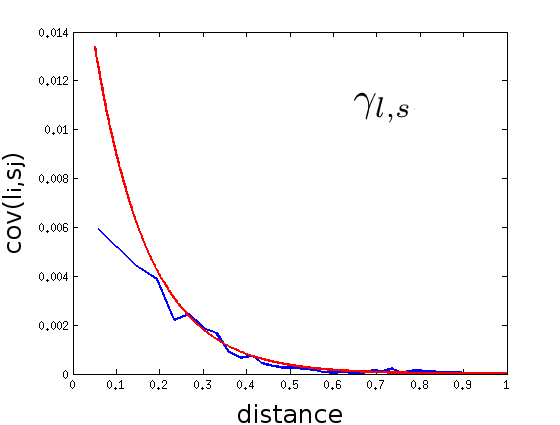}
		%\caption{}
	\end{subfigure}%
	\caption{Covariance as a function of patch appearance distance. (a) Covariance of detection scores $\gamma_{\scoresimbol,\scoresimbol}$; (b) Covariance of detection score and patch label $\gamma_{\labelsimbol,\scoresimbol}$.}
	\label{fig: Covariance}
\end{figure}

Figure~\ref{fig: Covariance} displays in blue the estimated $\gamma_{\scoresimbol,\scoresimbol}$ and $\gamma_{\labelsimbol,\scoresimbol}$ (using the data collected from SUN09 and equations~\eqref{eq:gamma_scores} and~\eqref{eq:gamma-score-and-label}). As can be seen, the estimated covariances are indeed monotonic descending functions of the patch distance. Best-fitting monotonic descending exponential functions are displayed in red. These analytic functions were used as the $\gamma$ functions in the CISS experiments described in Section~\ref{VOC results}.

\subsection{CISS Implementation Details}\label{implementaiton-details}
In this section we describe the details of the CISS implementation, when using the Fast R-CNN detector~\cite{girshickICCV15fastrcnn} as the base detector that provides the base-scores.
Fast R-CNN returns for each image a set of detections, each defined by a rectangle region, a class and a base-score.
In our implementation, anchors are regions whose base-scores were above $0.05$ and whose width and height exceeds 15 pixels. For evaluation purposes, other boxes provided by the detector are treated as anchors that have no supporters. Consequently, Eq.~\eqref{eq:MSE_solution} is applied on all input boxes, and all have comparable `CISS-scores'.

%We recursively divide the patches into four equal parts, and compute the distance between every two parts in the same quadrant, using the same descriptors and measure described above. The sub-part distances are combined with the full-patch distance.
%We compute second histogram maps for half the size in both width and height, and compute chi square distance between anchor sub-patches to the current map.
%This is the spatial component of the similarity.
%We continue to divide the windows until the sub-patches are to small to have meaningful content.
%After computing the full patch distance, and all the spatial component, we average the values, so each spatial level will have the same weight in the final calculations.\\

%\subsubsection{Efficiently locating supporters}\label{Searching}
To efficiently locate supporters, the color and textural features described above are extracted for all image locations and for all relevant scales using integral images. Given the anchor's dimensions $w_i\times h_i$, we search for supporters of dimensions between $0.8w_i\times 0.8h_i$ and $1.2w_i\times 1.2h_i$, having the same aspect ratio as $b_i$.
The chi-square distance from the anchor’s descriptors is then computed in parallel for each possible image location.
Given the distances between the anchor's descriptors and all possible locations for supporters, we select up to $n_{\text max}$ supporters with distance $d_{i,j}$ not exceeding 0.25. We also limit the choice of supporters so that they do not overlap with the anchor or with each other. Thus, we iteratively select the most similar (lowest feature-wise distance) supporter not overlapping with previous selections.
The search for supporters is run on a GeForce GTX TITAN X GPU.

\subsection{CISS Results on PASCAL VOC}\label{VOC results}
% intro
In this section we report results of CISS on the PASCAL VOC2007~\cite{pascal-voc-2007} benchmark detection challenge.
The Fast R-CNN detector~\cite{girshickICCV15fastrcnn} with its CaffeNet model is used both as a baseline for comparison and as the base detector providing the base scores.

\begin{figure*}[tbp]
	\centering
	\begin{subfigure}[b]{0.27\textwidth}\
		\includegraphics[width=\linewidth]{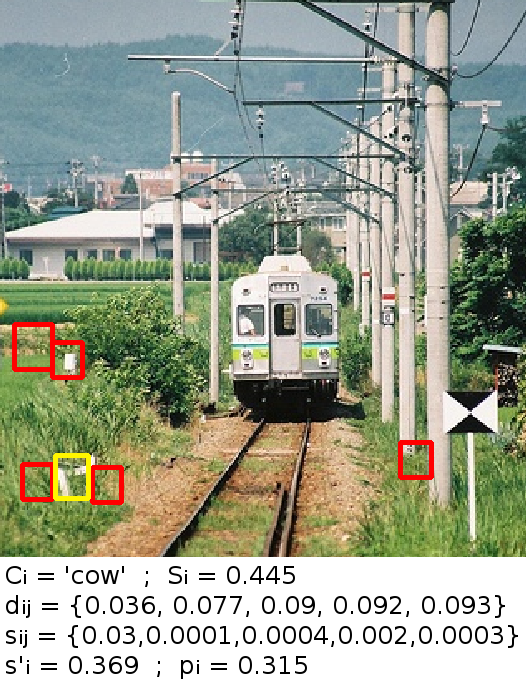}
			%img_1084_win_4742_redone.png}
		\caption{} \label{fig:similarity_examples_pascal_similarity_parts_4}
	\end{subfigure}%
	\hfill
	\begin{subfigure}[b]{0.23\textwidth}\
		\includegraphics[width=\linewidth]{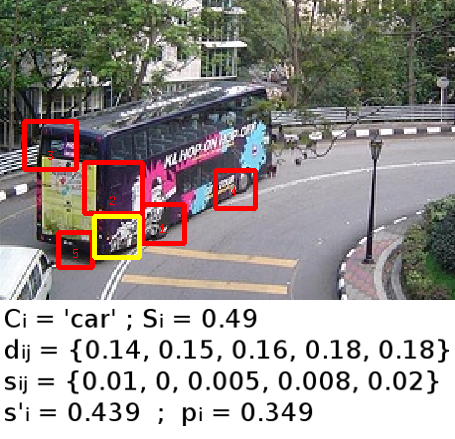}
				\caption{}
				\label{fig:similarity_examples_pascal_similarity_parts_1}
	\end{subfigure}%	
	\hfill		
				\begin{subfigure}[b]{0.25\textwidth}\
				\includegraphics[width=\linewidth]{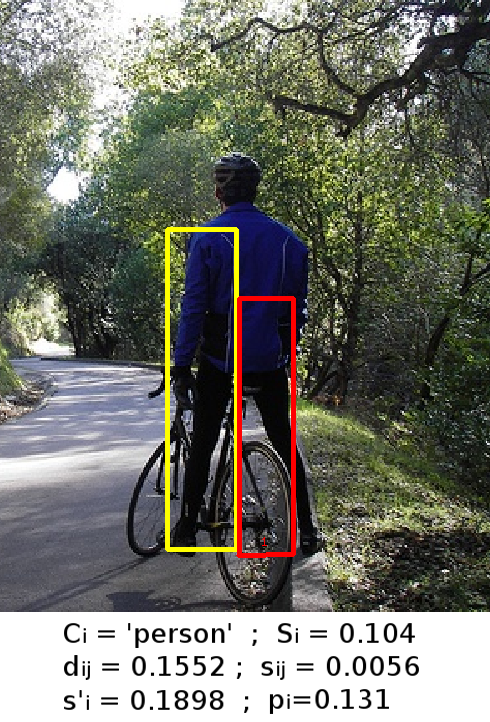}
				\caption{} \label{fig:similarity_examples_pascal_similarity_parts_2}
				\end{subfigure}%
		\hfill		
				\begin{subfigure}[b]{0.2\textwidth}\
				\includegraphics[width=\linewidth, height =6cm]{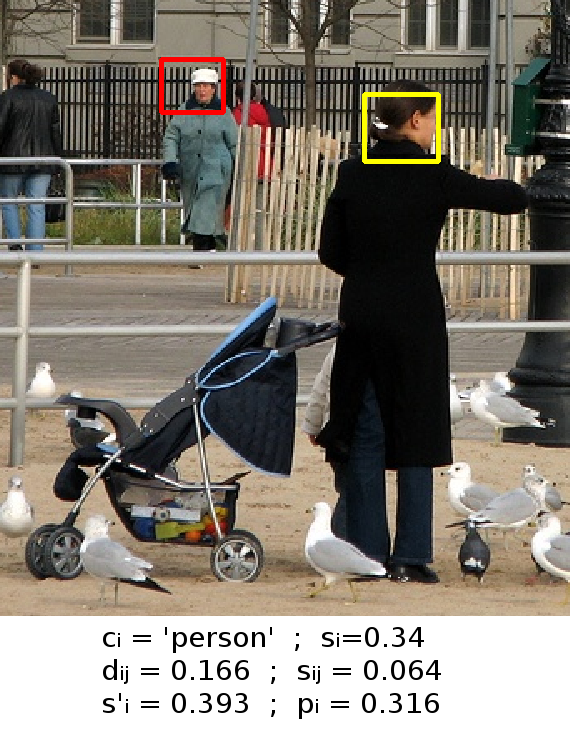}
				\caption{} \label{fig:similarity_examples_pascal_similarity_parts_3}
				\end{subfigure}%	
		\\
	\begin{subfigure}[b]{0.35\textwidth}\			
	\includegraphics[width=\linewidth]{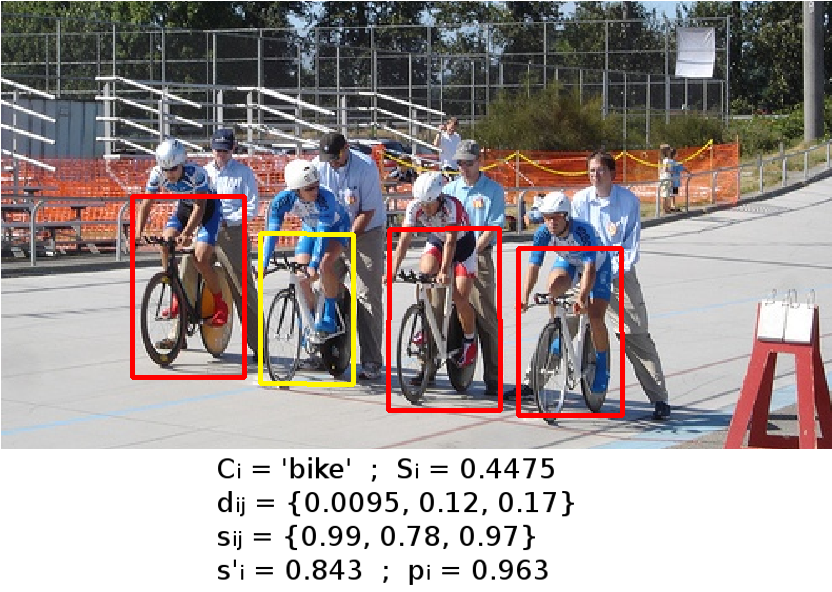}
		\caption{}\label{fig:similarity_examples_pascal_similarity_other_objects_1}
		\end{subfigure}%
		\hfill
		\begin{subfigure}[b]{0.36\textwidth}\
			\includegraphics[width=\linewidth]{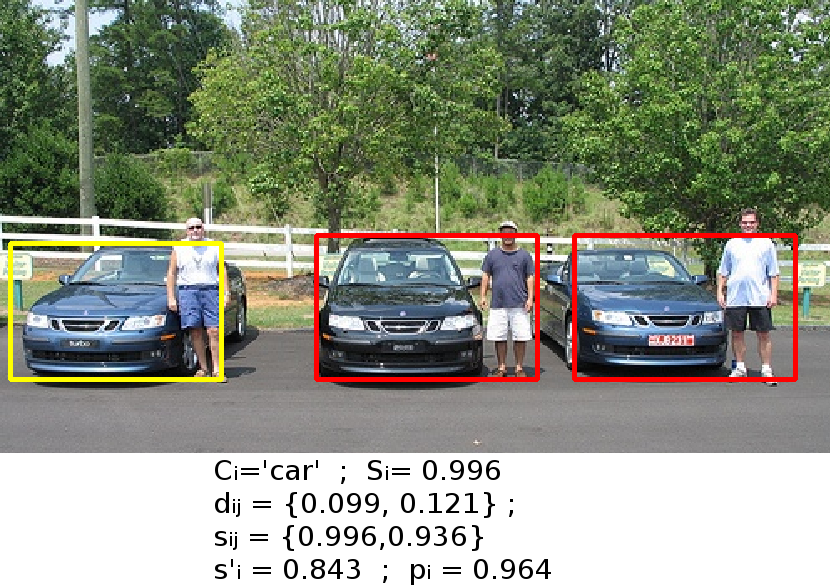}
			\caption{}\label{fig:similarity_examples_pascal_similarity_other_objects_2}
		\end{subfigure}%
			\hfill		
			\begin{subfigure}[b]{0.25\textwidth}\
					\includegraphics[width=\linewidth]{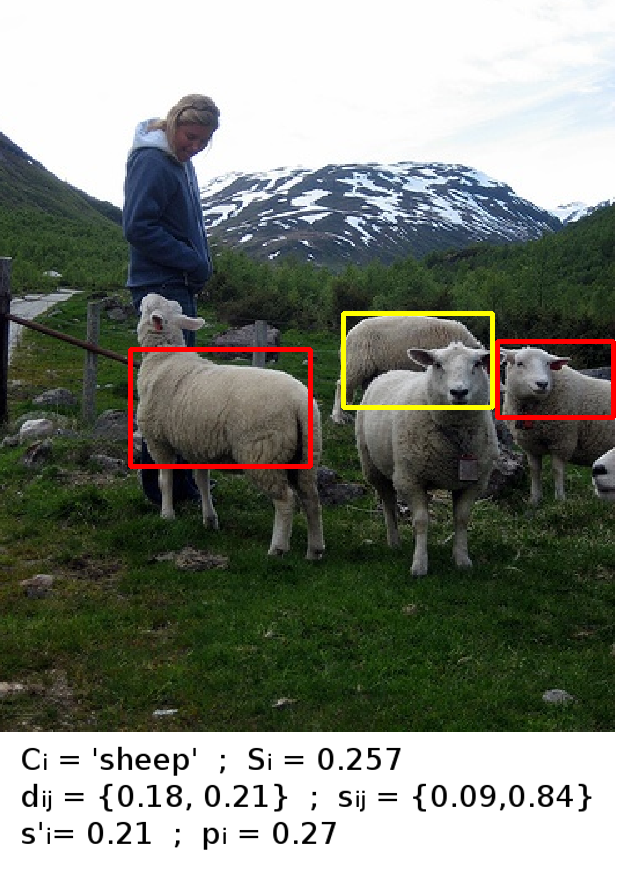}
					\caption{}\label{fig:similarity_examples_pascal_similarity_other_objects_4}
			\end{subfigure}%
%			\\
%			\begin{subfigure}[b]{0.32\textwidth}\
%				\phantom{}
%			\end{subfigure}%
	\caption{Some CISS results}
	\label{fig:similarity_examples_pascal}
\end{figure*}

% Examples in figures
Examples of CISS's results can be seen in Figure~\ref{fig:similarity_examples_pascal}.
The base-score $s_i$ and the CISS-score $p_i$ cannot be directly compared, so for each anchor, we also apply Eq.(~\ref{eq:MSE_solution}) as if it had no supporters. This revised base-score is annotated by ${s_i}'$ and is comparable with $p_i$. We therefore report $s_i'$ in the examples in Figure ~\ref{fig:similarity_examples_pascal}. Note that the transformation from $s_i$ to $s_i'$ is monotonic and preserves the order between base-scores (for each category).
The upper row of the Figure shows cases in which CISS reduced the score of background patches or object parts (reducing false alarms). The second row of Figure~\ref{fig:similarity_examples_pascal} shows cases where CISS increased the scores of objects due to their similarity to other objects in the scene (increasing detection rate).

%In Figure~\ref{fig:similarity_examples_pascal_similarity_parts_4}, for instance, we see a case where a background-cluttered patch was mistakenly detected as an object (sheep) by the base detector. In this case CISS found a number of similar patches with low base-scores and was able to reduce the estimated probability for it to contain a sheep. For objects that are commonly found in "groups" (e.g. a herd of cows or a bunch of bottles), the instances that score very high can improve the detection of low-scoring or partially occluded instances; See Figure~\ref{fig:similarity_examples_pascal_similarity_other_objects_3}.

\begin{figure}[tbp]
	%\centering
	\begin{subfigure}[b]{0.25\textwidth}
		\includegraphics[width=\textwidth]{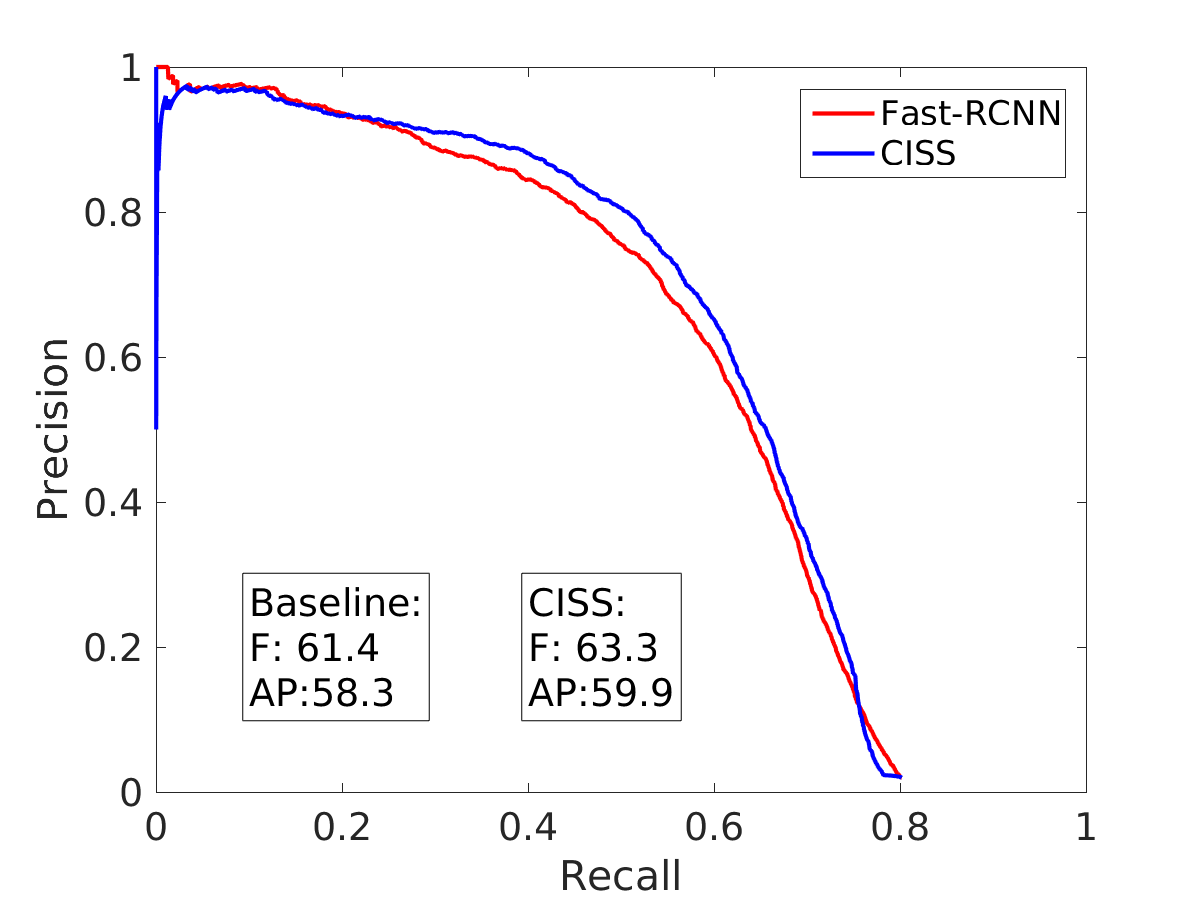}
		\caption{person}\label{fig:PR_cls1_all}
	\end{subfigure}%
	~
	\begin{subfigure}[b]{0.25\textwidth}
		\includegraphics[width=\textwidth]{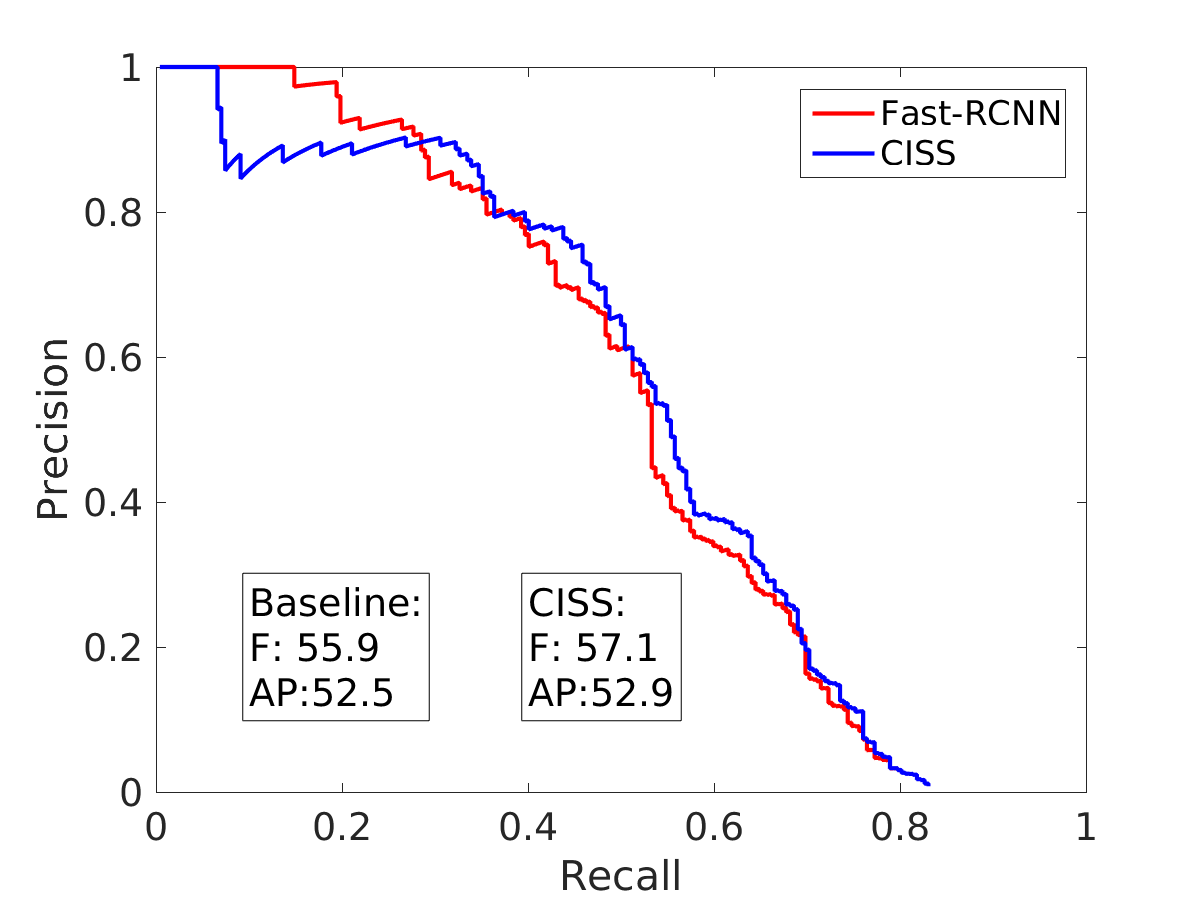}\caption{sheep}\label{fig:PR_cls2_all}
	\end{subfigure}%	
	\vfill
	\begin{subfigure}[b]{0.25\textwidth}
		\includegraphics[width=\textwidth]{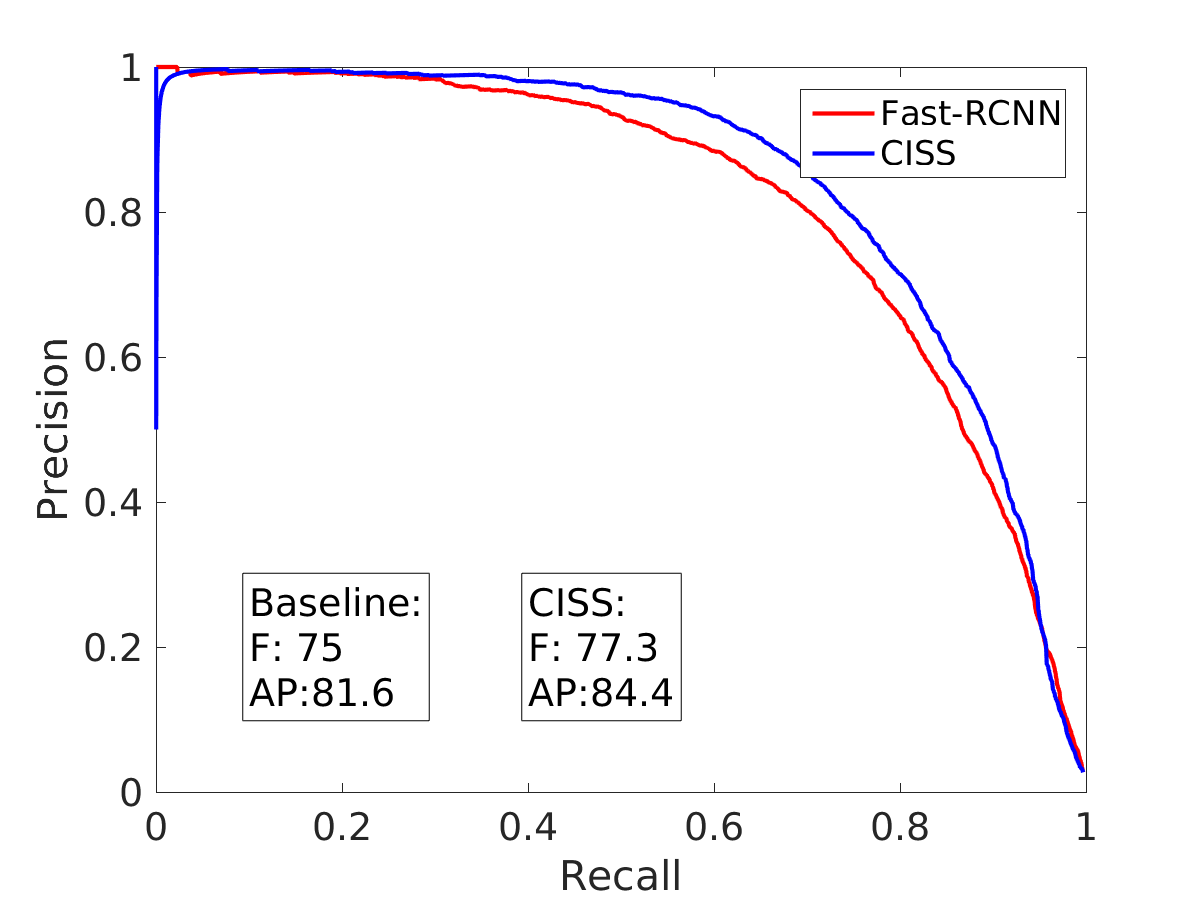}
		\caption{person}\label{fig:PR_cls1_ignore_locsim}
	\end{subfigure}%
	~
	\begin{subfigure}[b]{0.25\textwidth}
		\includegraphics[width=\textwidth]{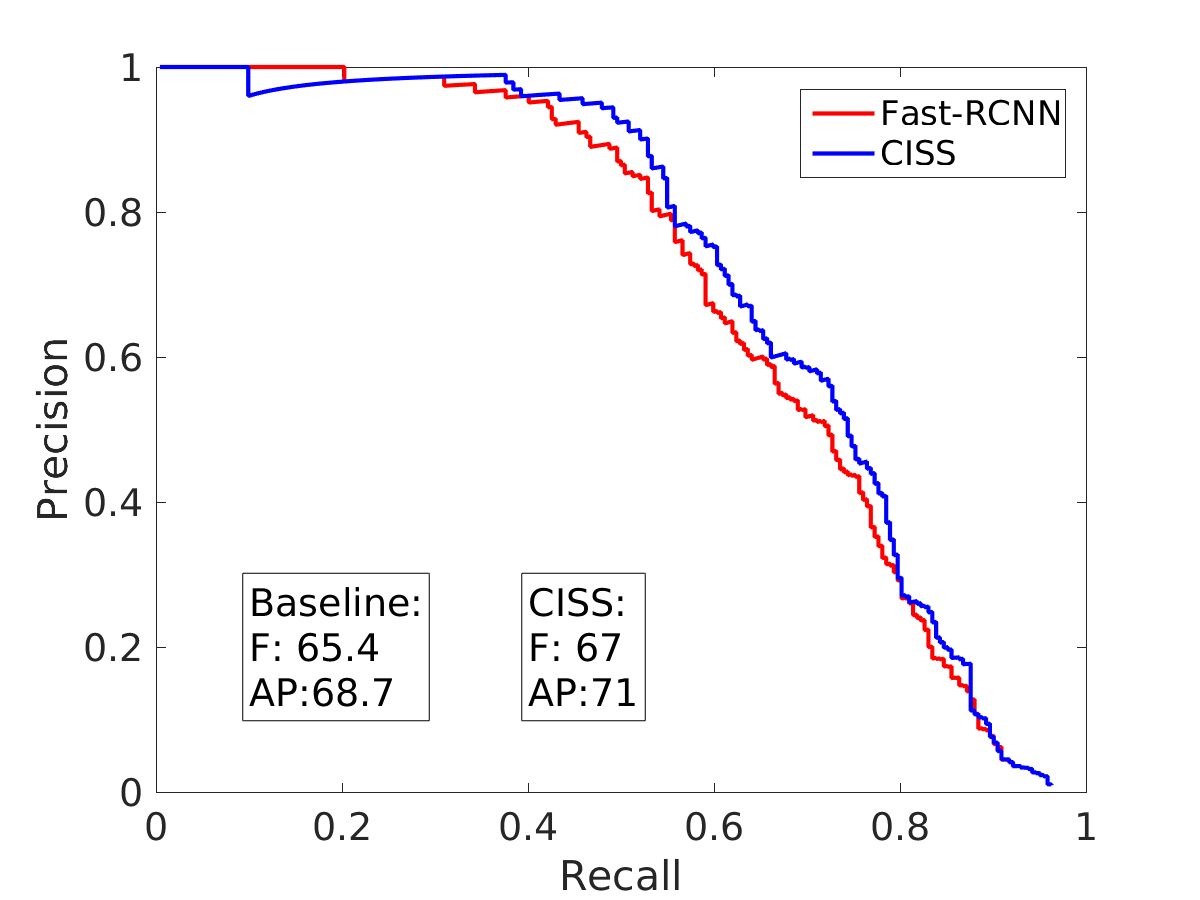}                				\caption{sheep}\label{fig:PR_cls2_ignore_locsim}
	\end{subfigure}%
	
	%		\caption{Precision-Recall curve - Ignoring localization and similar object errors}
	%\label{fig:Precision-Recall curves - ignore sim+loc}
	\caption{Precision-Recall curves. Top line - with all errors taken into account. Bottom line - when ignoring location and similar object errors}\label{fig:Precision-Recall curves}
\end{figure}

% precision in low recall vs. high recall
In order to evaluate the CISS algorithm, we compared the precision vs. recall (P-R) curves of Fast R-CNN and CISS for the 20 PASCAL categories.
See two examples in Figure~\ref{fig:PR_cls1_all} and~\ref{fig:PR_cls2_all}. As demonstrated, CISS's results are sometimes inferior for low recalls and usually improve precision for higher recalls. Note that for most applications, the low recall section of the P-R curve has no practical importance. 
% AP is not a good eval method
Yet, the most common evaluation measure for object detection is mean average precision (mAP), which is equal to the area under the P-R curve. This area averages the performance for low and high recall, providing the same weight for both. The F-score measure, on the other hand, compares the best performing point in the P-R curve of different detectors. We summarize the results for both criteria in Table~\ref{tab:all-results}. As can be seen CISS improves the F-score results for 17 of the 20 PASCAL categories, while preserving the F-score for all other categories but one. The mAP is also improved, for 11 of the classes.

\begin{table*}[bt]
	\centering	
	\resizebox{\textwidth}{!}{  		
		\begin{tabular}{*{22}{c}}
			\toprule
			& plane & bike & bird & boat & bottle& bus & car & cat & chair & cow & table & dog & horse & motorbike  & person & plant & sheep & sofa & train & tv & Avg\\
			\midrule	
			\multicolumn{1}{c}{mAP}\\
		Fast R-CNN & 68.4&	73.2&	53.8&	42.7&	22.5&	\textbf{70.6}&	72.0&	\textbf{74.0}&	29.2&	64.2&	60.8&	\textbf{63.6}&	75.7&	\textbf{69.5}&	58.3&	\textbf{23.1}&	52.5&	\textbf{56.6}&	\textbf{69.9}&	\textbf{59.2}& 58\\
		Fast R-CNN + CISS & \textbf{68.5}& \textbf{73.6}&	\textbf{54.6}&	\textbf{44.0}&	\textbf{23.4}&	70.2&	\textbf{72.2}&	73.9&	\textbf{31.4}&	\textbf{64.5}&	60.8&	63.4&	\textbf{76.1}&	68.7&	\textbf{59.9}&	22.3&	\textbf{52.9}&	55.7&	69.8&	56.6& \textbf{58.1}\\
			\midrule
			\multicolumn{1}{c}{F-score}\\	
		Fast R-CNN & 68.3&	72.2&	57.4	& 49.4&	30.3&	\textbf{70.6}&	73.3&	74.5&	36.9&	64.2&	63.9&	64.8&	74.4&	69.9&	61.4&	30.5&	55.9&	59.1&	\textbf{71.9}&	59.5	& 60.4 \\
		Fast R-CNN + CISS & \textbf{69.1}&	\textbf{72.7}& \textbf{58.4}&	\textbf{50.3}& \textbf{31.1}&	70.2&	\textbf{73.6}&	74.5&	\textbf{38.0}&	\textbf{66.0}&	\textbf{64.1}& 64.8&	\textbf{74.5}&	\textbf{70.2}&	\textbf{63.3}&	\textbf{31.8}&	\textbf{57.1}&	\textbf{59.2}&	71.5&	\textbf{60.1}& \textbf{61}\\
			\bottomrule
		\end{tabular}
	}
	\caption{Detection results when considering all errors for PASCAL VOC 2007 test}
	\label{tab:all-results}
\end{table*}

\begin{table*}[bt]
	\centering	
	\resizebox{\textwidth}{!}{  		
		\begin{tabular}{*{22}{c}}
			\toprule
			& plane & bike & bird & boat & bottle& bus & car & cat & chair & cow & table & dog & horse & motorbike  & person & plant & sheep & sofa & train & tv & Avg\\
			\midrule	
			\multicolumn{1}{c}{mAP}\\
			Fast R-CNN& 82.2&	86.4&	67.0&	68.6&	33.1&	81.2&	83.0&	92.8&	48.7&	87.6&	77.8&	91.6&	91.9&	86.1&	81.6&	42.2&	68.8&	77.7&	89.1&	\textbf{62.7} & 75.1\\
			Fast R-CNN + CISS& \textbf{82.8}& \textbf{86.8}&	\textbf{70.5}&	\textbf{69.2}&	\textbf{35.7}&	\textbf{81.5}&	\textbf{84.5}&	\textbf{93.0}&	\textbf{51.6}&	\textbf{88.5}&	77.8&	\textbf{92.0}&	\textbf{92.1}&	\textbf{86.4}&	\textbf{84.4}&	\textbf{43.7}&	\textbf{71.0}&	\textbf{77.8}&	\textbf{89.6}&	60.6& \textbf{76} \\
			\midrule
			\multicolumn{1}{c}{F-score}\\	
			Fast R-CNN & 80.7&	80.7&	67.3&	65.5&	36.8&	\textbf{78.4}&	79.1&	88.8&	50.3&	81.5&	73.6&	86.7&	87.9&	80.9&	75.0&	44.2&	65.4&	\textbf{76.1}&	84.7&	61.7 & 72.3 \\
			Fast R-CNN + CISS &  \textbf{81.3}& \textbf{81.6}&	\textbf{68.3}& \textbf{66.5}& \textbf{38.0} & 78.1& \textbf{80.1}& \textbf{88.9}& \textbf{51.6}& \textbf{82.9}& \textbf{73.8}& \textbf{87.3}& 87.9& \textbf{81.2}& \textbf{77.3}& \textbf{45.7}& \textbf{67.0}& 76.0& 84.7& \textbf{62.5}& \textbf{73} \\
			\bottomrule
		\end{tabular}
	}
	\caption{Detection results when ignoring location and similar object error for PASCAL VOC 2007 test}
	\label{tab:ignor-loc-sim-results}
\end{table*}

In previous works~\cite{hoiem2012diagnosing, redondo2016pose} several other drawbacks have been demonstrated for the mAP measure, and alternative measures have been proposed. It was suggested in~\cite{hoiem2012diagnosing} that localization and `similar-object' errors are less important than other errors. We followed this observation and tested how CISS performs when ignoring the localization and `similar-object' errors. As can be seen in Figure~\ref{fig:PR_cls1_ignore_locsim} and~\ref{fig:PR_cls2_ignore_locsim}, CISS's disadvantage for low recalls is significantly reduced, while the advantage for high recalls is maintained. See Table~\ref{tab:ignor-loc-sim-results} for mAP and F-scores of CISS vs. Fast R-CNN when only considering the more crucial errors.

\begin{figure*}[bpt]
	\centering		
	\hfill	
	\begin{subfigure}[b]{0.3\textwidth}\
		\includegraphics[width=\linewidth]{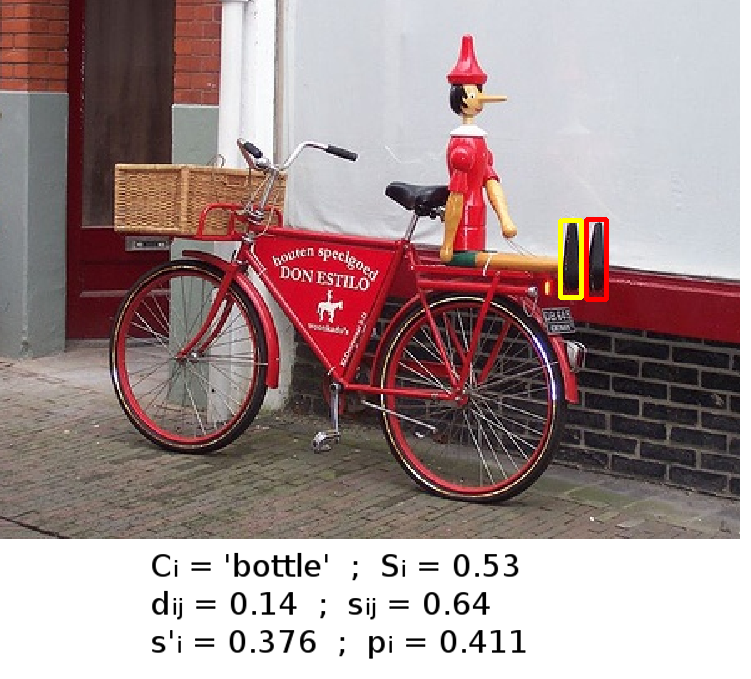}
		\caption{}\label{fig:Failure_cases_1}
	\end{subfigure}%
	\hfill	
		\begin{subfigure}[b]{0.35\textwidth}\
			\includegraphics[width=\linewidth]{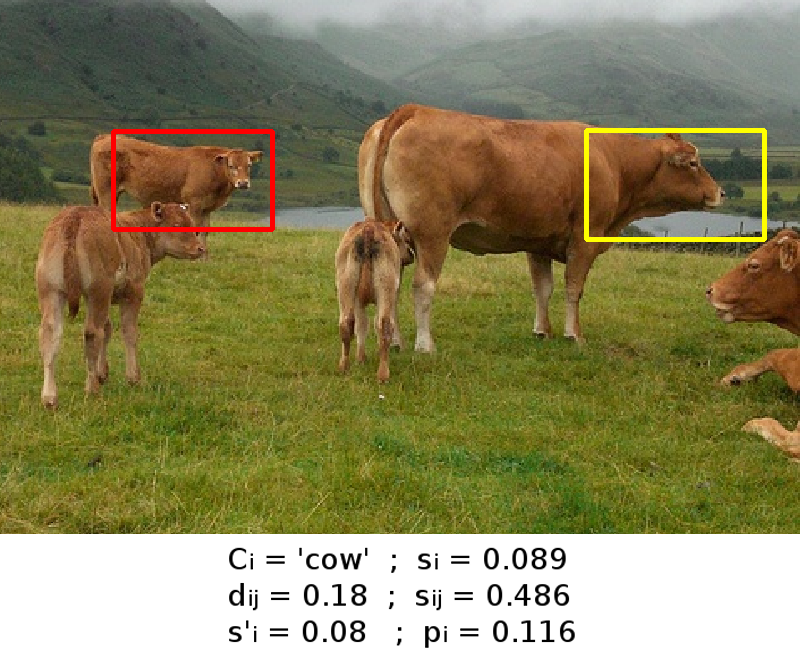}
			\caption{}\label{fig:Failure_cases_2}
		\end{subfigure}%
	\hfill		
			\begin{subfigure}[b]{0.3\textwidth}\
				\includegraphics[width=\linewidth]{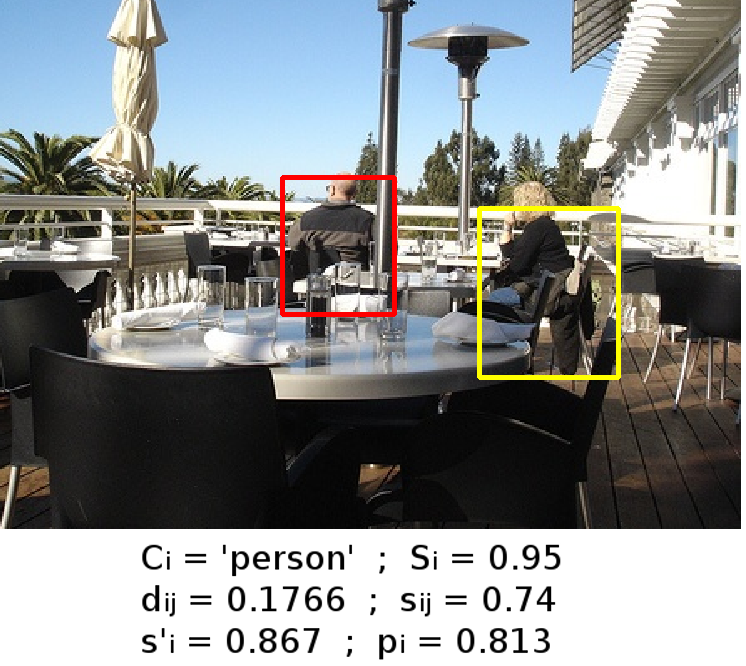}				
				\caption{}\label{fig:Failure_cases_3} % \label{fig:similarity_examples_pascal_true_occ}
			\end{subfigure}% 
			\\
			\begin{subfigure}[b]{0.32\textwidth}\
			\phantom{}
				\end{subfigure}% 
	\caption{Failure cases}
	\label{fig:Failure_cases}
\end{figure*}

\subsection{Failure Cases and Discussion}\label{discussion}
Figure~\ref{fig:Failure_cases} demonstrate some failure cases. In Figure~\ref{fig:Failure_cases_1} we demonstrate a case where the supporter is mistakenly classified by the base detector with a high base-score (to be an instance of the class `bottle'), which leads incorrectly to an increase of the anchor's score.
In Figure~\ref{fig:Failure_cases_2} we demonstrate a case where an object part is found to be similar to a full object. As a result, CISS increased the score of the part.
In Figure~\ref{fig:Failure_cases_3} we demonstrate a case where the anchor and supporter are both of the same class (a person), but the base detector assigns a low base-score to the supporter. As a result CISS decreased the score of the anchor.

Several extensions to consider for avoiding such failures are described in Section~\ref{Conclusion}. Nonetheless, it should be noted that the CISS algorithm improves detection accuracy more often than decreasing it. When the goal is a better operating point, incorporating CISS as a post-detection step is worthwhile.

\section{Conclusion} \label{Conclusion}
% summery
In this work, we explore inner scene similarities as a contextual cue for enhancing object detection performance. Our method relates to a psychophysical study which suggested that the human visual system perceptually groups similar image regions and then estimates their identity simultaneously. Statistical analysis confirms that patches sharing visual properties also share semantic identities. We have shown that the suggested CISS algorithm improves the operating point for the Fast R-CNN detector by reducing the number of false alarms and enhancing detection of partly occluded objects. 
Interestingly, we used a similarity measure that is based on descriptors that are variant in pose and illumination, an approach which goes against the common computer vision wisdom. Yet comparing patches in this manner allows us to use the increased similarity between the same objects in the same image, which is neglected by the classifiers.

\begin{figure*}[bpt]
	\centering
	\begin{subfigure}[b]{0.27\textwidth}
		\includegraphics[height=\textwidth]{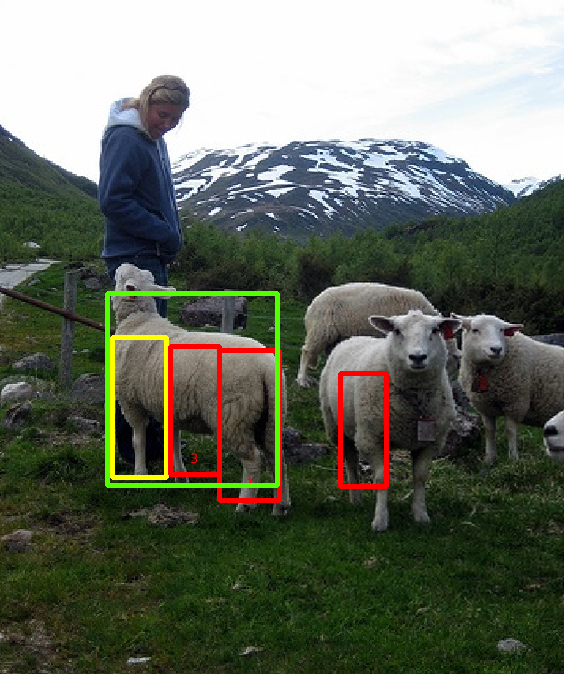}
		\caption{}
		%\caption{Object part scores are enhanced (unjustifiably) due to other parts. If we knew they were parts of a full object, this could be avoided.}
		\label{fig:parts-example}
	\end{subfigure}%
	\hfill
	\begin{subfigure}[b]{0.3\textwidth}
		\includegraphics[width=1.1\textwidth]{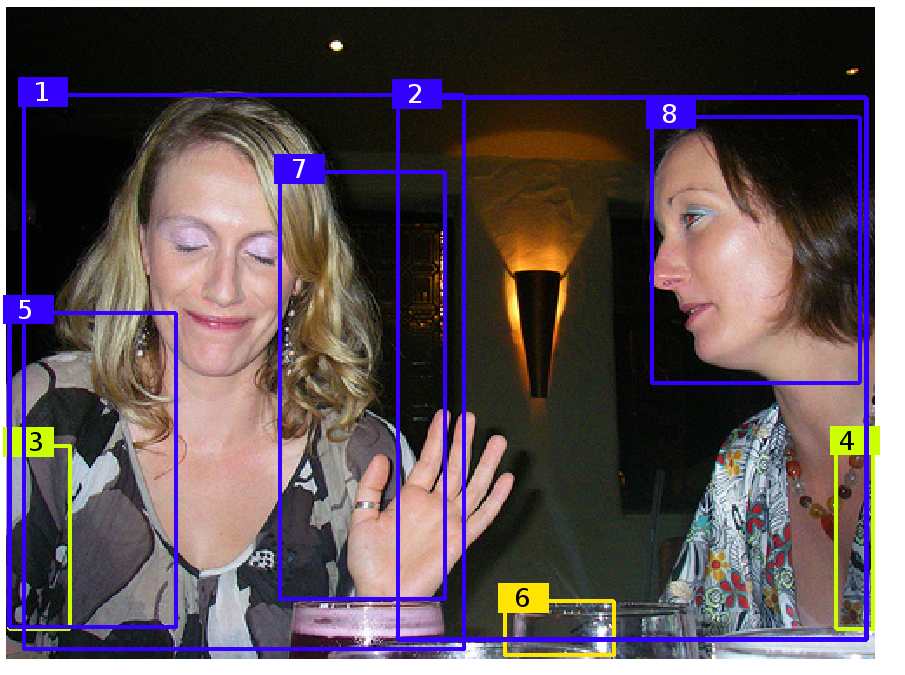}	\caption{}
		%\caption{Best 8 detections, Fast R-CNN}
		\label{fig:nms-results-frcnn}
	\end{subfigure}%
	\hfill
	\begin{subfigure}[b]{0.3\textwidth}
		\includegraphics[width=1.1\textwidth]{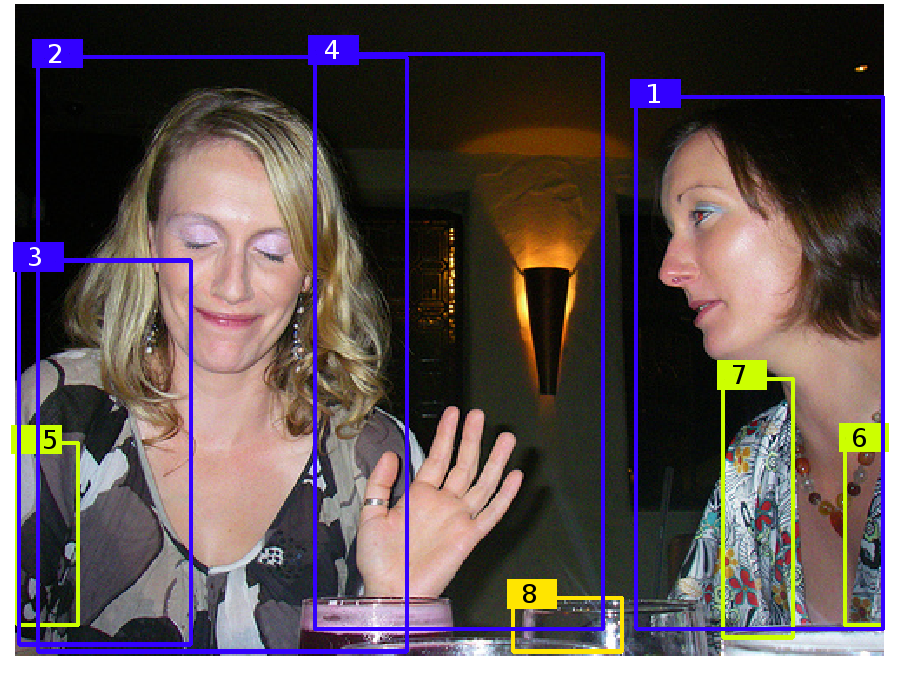}
		\caption{}
		%\caption{Best 8 detections, NMS after CISS}
		\label{fig:nms-results-ciss}
	\end{subfigure}% 			
		\caption{(a) Object part scores are enhanced (unjustifiably) due to other parts. If we knew they were parts of a full object, this could be avoided.
			(b)-(c) Illustration of the benefit of applying CISS before NMS, resulting in a better localization of the woman on the right.}
		\label{fig:nms-and-parts-example}
\end{figure*}

In this work, we showed how the inner-scene similarity cue can contribute by itself, but combining it with other contextual cues can be helpful.
While CISS uses the MMSE linear estimator, better results may be possible if a general estimator is learned.
Detectors often fail to confuse object-parts and full-objects. This follows the desire to detect partly occluded objects, for which only a part is visible. When the detector mistakenly highly scores a part of a fully visible object, CISS can make things worse by increasing the score of more similar parts (see Figure~\ref{fig:parts-example}). Treating differently anchors that are contained inside other highly scored anchor (and therefore likely to be parts) may help prevent undesired affects of CISS.
An interesting route to follow is the exploitation of inner-scene similarities for an intelligent non-maximum suppression (NMS) process.
In order to avoid multiple detections of one object instance, NMS is usually applied on the output of the detector in a greedy process.
% the solution using CISS
The CISS algorithm can produce better localization by changing the scores of all detection windows before applying NMS.
See Figure~\ref{fig:nms-results-frcnn} and~\ref{fig:nms-results-ciss}.

{\small
\bibliographystyle{ieee}
\bibliography{CISS_bib}
}

\end{document}